\documentclass[conference]{IEEEtran}
\IEEEoverridecommandlockouts
\usepackage{cite}
\usepackage{amsmath,amssymb,amsfonts}
\usepackage{algorithmic}
\usepackage{graphicx}
\usepackage{textcomp}
\usepackage{xcolor}
\def\BibTeX{{\rm B\kern-.05em{\sc i\kern-.025em b}\kern-.08em
    T\kern-.1667em\lower.7ex\hbox{E}\kern-.125emX}}

\usepackage{booktabs}
\usepackage{caption}
\usepackage{subcaption}
\usepackage{color}
\usepackage[ruled,linesnumbered]{algorithm2e}
\usepackage{tabularx}
\usepackage[misc]{ifsym}

\begin{document}
\bstctlcite{MyBSTcontrol}

\title{Exploring Unknown States with Action Balance
\thanks{}
}

\author{\IEEEauthorblockN{Song Yan \textsuperscript{\Letter}}
\IEEEauthorblockA{
\textit{Fuxi AI Lab in Netease}\\
Hangzhou, China \\
songyan@corp.netease.com}
\and
\IEEEauthorblockN{Chen Yingfeng \textsuperscript{\Letter}}
\IEEEauthorblockA{
\textit{Fuxi AI Lab in Netease}\\
Hangzhou, China \\
chenyingfeng1@corp.netease.com}
\and
\IEEEauthorblockN{Hu Yujing}
\IEEEauthorblockA{
\textit{Fuxi AI Lab in Netease}\\
Hangzhou, China \\
huyujing@corp.netease.com}
\and
\IEEEauthorblockN{Fan Changjie}
\IEEEauthorblockA{
\textit{Fuxi AI Lab in Netease}\\
Hangzhou, China \\
fanchangjie@corp.netease.com}
}

\IEEEpubid{\begin{minipage}{\textwidth}\ \\[12pt]
978-1-7281-4533-4/20/\$31.00 \copyright 2020 IEEE
\end{minipage}}

\maketitle

\begin{abstract}
Exploration is a key problem in reinforcement learning. Recently bonus-based methods have achieved considerable successes in environments where exploration is difficult such as Montezuma's Revenge, which assign additional bonuses (e.g., intrinsic rewards) to guide the agent to rarely visited states. Since the bonus is calculated according to the novelty of the next state after performing an action, we call such methods as the \textit{next-state bonus} methods. However, the \textit{next-state bonus} methods force the agent to pay overmuch attention in exploring known states and ignore finding unknown states since the exploration is driven by the next state already visited, which may slow the pace of finding reward in some environments. In this paper, we focus on improving the effectiveness of finding unknown states and propose \textit{action balance exploration}, which balances the frequency of selecting each action at a given state and can be treated as an extension of upper confidence bound (UCB) to deep reinforcement learning. Moreover, we propose \textit{action balance RND} that combines the \textit{next-state bonus} methods (e.g., random network distillation exploration, RND) and our \textit{action balance exploration} to take advantage of both sides. The experiments on the grid world and Atari games demonstrate \textit{action balance exploration} has a better capability in finding unknown states and can improve the performance of RND in some hard exploration environments respectively.
\end{abstract}


\section{Introduction}
Reinforcement learning methods are aimed at learning policies that maximize the cumulative reward. The state-of-the-art RL algorithms such as DQN \cite{mnih_human-level_2015}, PPO \cite{schulman_proximal_2017} work well in dense-reward environments but tend to fail when the environment has sparse rewards, e.g. Montezuma's Revenge. This is because with the immediate rewards of most state-action pairs being zero, there is little useful information for updating policy. Reward shaping \cite{ng_policy_1999} is one solution that introduces human expertise and converts the original sparse problem to a dense one. However, this method is not universal, and transforming human knowledge into numeric rewards is usually complicated in real tasks. 

\begin{figure}
    \centering
    \begin{subfigure}{0.22\textwidth}
        \centering
        \includegraphics[ width=\textwidth]{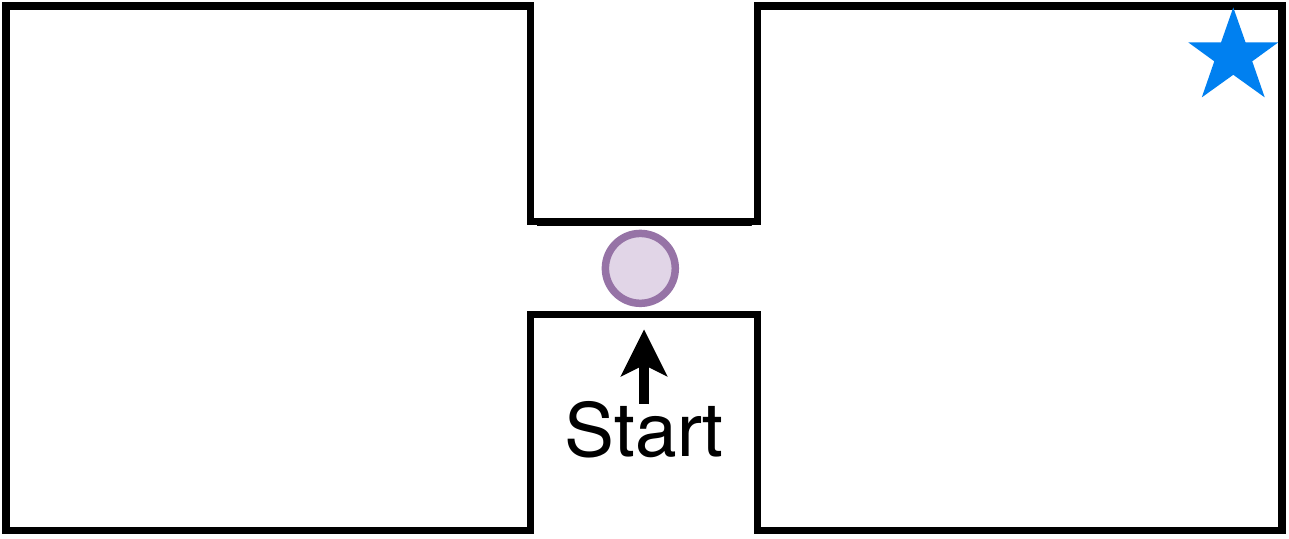}
        \caption{}
        \label{fig:nsb_example_a}
    \end{subfigure}
    \hspace{.08in}
    \begin{subfigure}{0.22\textwidth}
        \centering
        \includegraphics[ width=\textwidth]{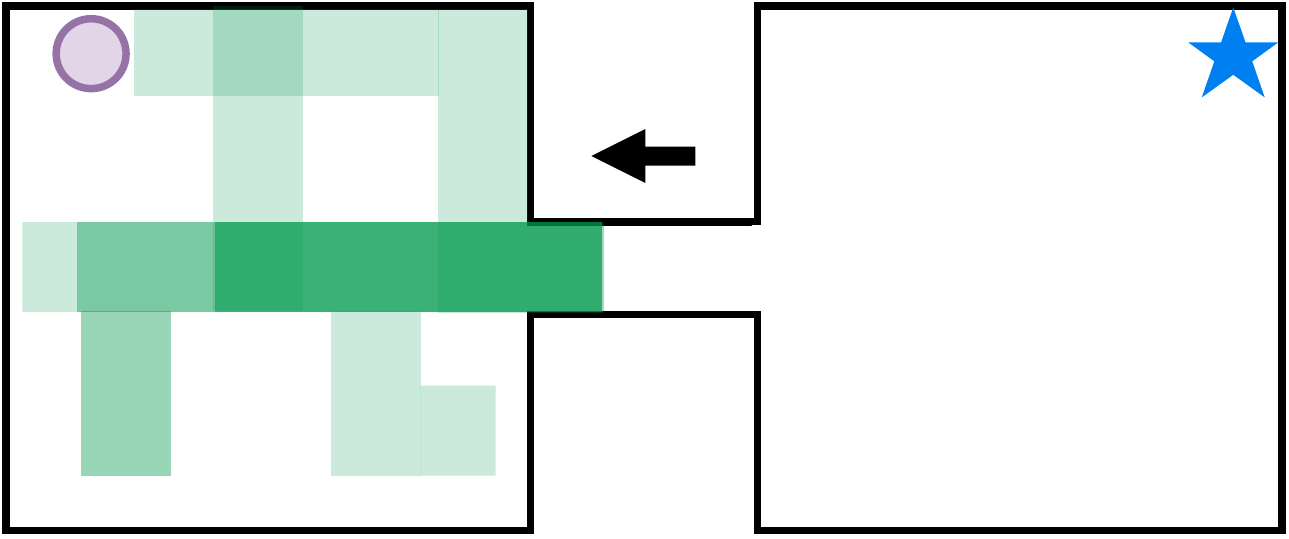}
        \caption{}
        \label{fig:nsb_example_b}
    \end{subfigure}
    \begin{subfigure}{0.22\textwidth}
        \centering
        \includegraphics[ width=\textwidth]{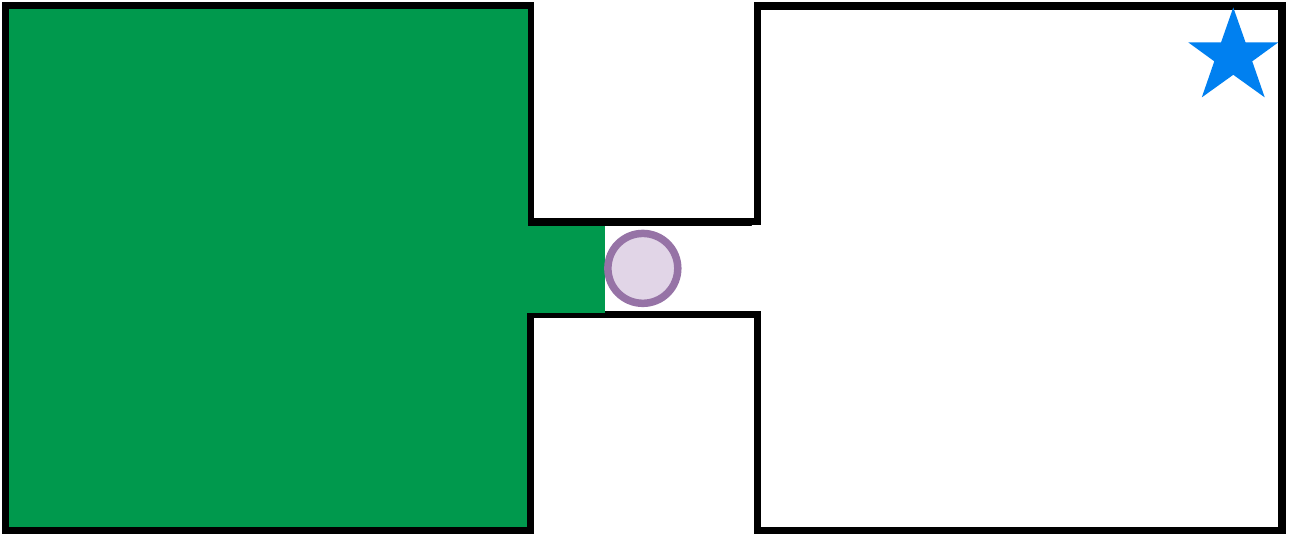}
        \caption{}
        \label{fig:nsb_example_c}
    \end{subfigure}
    \hspace{.08in}
    \begin{subfigure}{0.22\textwidth}
        \centering
        \includegraphics[ width=\textwidth]{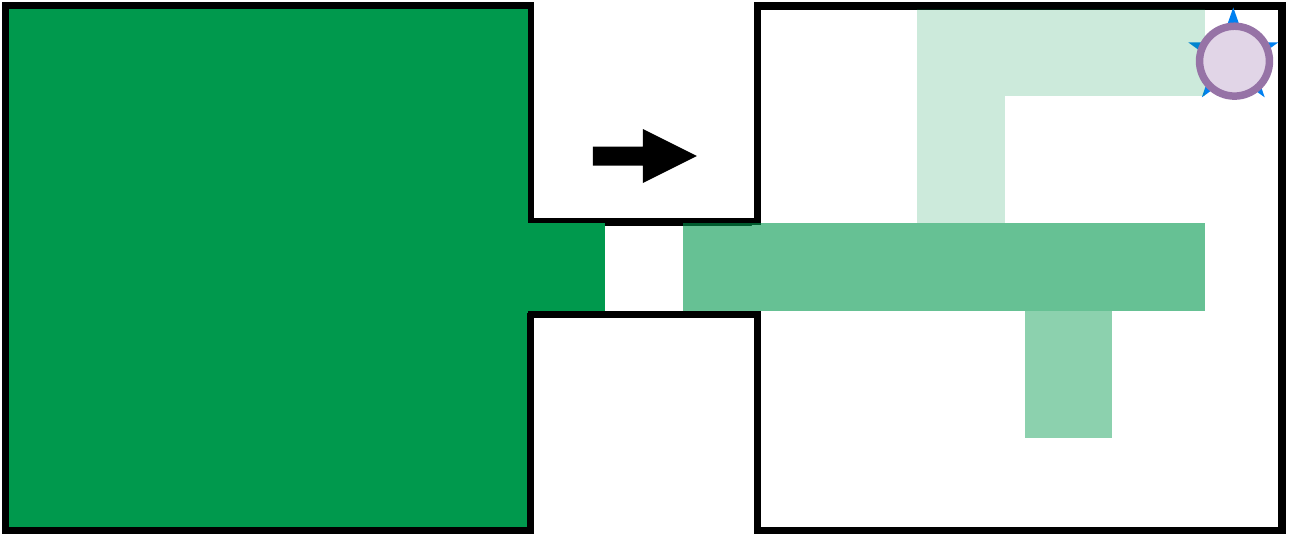}
        \caption{}
        \label{fig:nsb_example_d}
    \end{subfigure}

    \caption{The example of an agent dominated by the \textit{next-state bonus} method. The circle represents agent and the star represents treasure. Green indicates the visited area, the more time visited, the darker colored. a) Agent stands in the middle of the corridor at the start of each episode. b) Agent steps left by chance at first and continuously going left in the future (shallow green). c) The left area being explored exhaustedly (deep green). d) Agent decides to go right by chance and finally find the treasure.}
    \label{fig:nsb_example}
\end{figure}

Recently bonus based exploration methods have achieved great success in video games \cite{burda_exploration_2019,pathak_curiosity-driven_2017}. They use the next states obtained by performing some actions to generate intrinsic rewards (bonuses) and combine it with the external rewards to make the environment rewards dense. Since the bonus is calculated by the next state, we call them \textit{next-state bonus} methods for short in the following sections. However, the \textit{next-state bonus} force the agent to pay overmuch attention in exploring known states, but ignore finding unknown states since it takes effect by influencing the external rewards (details are described in Section \ref{sec:next-state-introduce}).

An illustration of the drawback in the \textit{next-state bonus} methods is shown in Figure \ref{fig:nsb_example}. Specifically, the action selection is completely based on the \textit{next-state bonus} and no other exploration strategies that exist, including $\epsilon-greedy$ and sampling actions from policy distribution. As shown in Figure \ref{fig:nsb_example_a}, two areas connected by a narrow corridor and a treasure is placed in the right area. An agent is born in the middle of the corridor and decides which area to explore at the beginning of each episode. At the very first step, since there is no transition information for the agent to calculate an additional bonus, which based on the next state, and the additional rewards of all states are equal to zero, it will decide which direction to go randomly. Assuming the agent chooses to go left by chance (Figure \ref{fig:nsb_example_b}), now transitions that all belong to the left area can be collected and additional non-zero bonuses will be assigned to the next states in the transitions, which makes the action to go left have a higher reward. At the start of the following episodes, the agent will continuously go left because of the higher reward and ignore the right area. When the left area is fully explored and all bonuses belong to it decay to zero (Figure \ref{fig:nsb_example_c}), the agent will face the same situation as the first step of the initial episode and select a direction randomly. Only until then, the agent may go right and find the treasure after many tries (Figure \ref{fig:nsb_example_d}).


This example shows how the agent pay overmuch attention in exploring known states (the left area) and ignore finding unknown states (the right area), which slows the pace of finding the treasure. In order to reduce the adverse effects that the \textit{next-state bonus} brings and stop going left all the time in Figure \ref{fig:nsb_example_b}, one solution is carrying out explorations when select action in $s_t$ directly, instead of regarding the bonus of $s_{t+1}$. For example, using $\epsilon-greedy$ exploration to select random actions with a certain probability will make it possible to go right at the start of each episode even if the left action has a higher reward (bonus) in Figure \ref{fig:nsb_example_b}. However, it is inefficient to rely entirely on fortunate behaviors.

Given these points, we propose a new exploration method, called \textit{action balance exploration}, that concentrates on finding unknown states. The main idea is to balance the frequency of selecting each action and it can be treated as an extension of upper confidence bound (UCB, \cite{auer_finite-time_2002}) to deep reinforcement learning. Specifically, in a given state, we record the frequency of selecting each action by using a \textit{random network distillation module} and generate a bonus for each action. Then, the action bonus vector will be combined with the policy $\pi$ to directly raise the probabilities of actions that are not often chosen. For an agent that uses the \textit{next-state bonus} methods, the \textit{action balance exploration} can avoid it from paying excessive attention to individual actions and improve the ability to find unknown states. For example, in the same situation as Figure \ref{fig:nsb_example_b}, the \textit{action balance exploration} will give a high priority to go right when standing in the middle of the corridor because the right action has a lower frequency of selecting than the left action. Moreover, our \textit{action balance exploration} method can be combined with the \textit{next-state bonus} methods (e.g., random network distillation exploration, RND) in a convenient way and take advantage of both sides. 

In this paper, we also combine our \textit{action balance exploration} method with RND and propose a novel exploration method, action balance RND, which can find unknown states more efficiently and simultaneously guide agents to visit unfamiliar states more frequently. We first test the action balance RND in a grid world that is the complete absence of rewards. The result shows that the action balance RND outperforms RND all through and overcomes the random baseline with about 2.5 times faster, which means the \textit{action balance exploration} improves the capability in finding unknown states of RND. Also, the action balance RND covers about 15 more grids, which is about 3\% higher in relative increase rate than RND at last. Second, in the environment of reaching goals, the action balance RND obtains the lowest trajectory length, which is about 1.23 times smaller on average than RND. Finally, we demonstrate that our \textit{action balance exploration} can improve the real performance of RND in some hard exploration Atari games.

\section{RELATED WORK}

Count-based methods
\cite{brafman_r-max_2001,kearns_near-optimal_2002} have a long research line, which use the novelty of states as an intrinsic bonus to guide exploration. Count-based methods are easy to implement and efficient in tabular environments, but they are not applicable to large scale problems once the state space is not enumerable. To solve this problem, many improvement schemes are proposed. TRPO-AE-hash \cite{tang_exploration:_2017} uses SimHash \cite{charikar_similarity_2002} to hash the state space. Although it can decrease the state space to some extent, it relies on the design of the hash algorithm.
DDQN-PC \cite{bellemare_unifying_2016}, A3C+ \cite{bellemare_unifying_2016}, DQN-PixelCNN \cite{ostrovski_count-based_2017}, and $\phi$-EB \cite{martin_count-based_2017} adopt density models \cite{oord_conditional_2016} to measure the visited time of states.   


ICM \cite{pathak_curiosity-driven_2017} and RND \cite{burda_exploration_2019} use the prediction error of supervised learning to measure state novelty since novel states are expected to have higher prediction errors due to the lesser training. \cite{savinov_episodic_2019} solves the 'noisy-TV' problem of ICM by introducing into a memory buffer. \cite{badia_never_2020} employs an additional memory buffer which saves the episodic states to calculate intrinsic rewards. \cite{strehl_theoretical_2005} theoretically analyzes the feasibility of using the number of $(s, a)$ pair occurs to estimate the upper confidence interval on the mean reward of the pair, which can be used to guide the exploration. Although many different methods for calculating intrinsic bonus \cite{taiga_benchmarking_2019, aubret_survey_2019} exist, the bonus usually takes effect by influencing the external reward from the environment. In contrast, UCB1 \cite{auer_finite-time_2002} records the frequency of selecting each action and gives high priorities to the actions that are not often selected, which is widely used in the tree-based search \cite{kocsis_bandit_2006, silver_mastering_2017}, but not suitable for innumerable state. DQN-UCB \cite{jin_is_2018} proves the validity of using UCB to perform exploration in Q-learning. Go-Explore \cite{ecoffet_go-explore:_2019} used a search-based algorithm to solve hard exploration problems.

Entropy-based exploration methods use a different way to maintain a diversity policy. \cite{mnih_asynchronous_2016} calculates the entropy of policy and adds it to the loss function as a regularization. This is easy to implement but only has a limit effect. Because it uses no extra information. \cite{hazan_provably_2019} provided a way to find a policy that maximizes the entropy in state space. The method works well in tabular setting environments, but hard to scale up in large state space. \cite{mei_principled_2019} provided a policy optimization strategy (Exploratory Conservative Policy Optimization, ECPO) that conducts maximum entropy exploration by changing the gradient estimator at each updating. The changed gradient makes it not only maximize the expected reward but also try to search for a policy with large entropy nearby. However, the computation relies on collected samples.

\section{METHOD}
\label{sec:method}
Consider an agent interacting with an environment. At each time step $t$, the agent obtained an observation $\mathbf{s}_t \in\mathcal{O}$ from the environment and samples an action with a policy $\pi(\mathbf{s}_t)$. After taking that action in the environment, the agent receives a scalar reward $r_t$, the new observation $\mathbf{s}_{t+1}$, and a terminal signal. The goal of the agent is to maximize the expected discounted sum of rewards $R=\sum_t{{\gamma^t}r_t}$. An environment that is hard to explore usually means the rewards are sparse (most of $r_t$ are zeros) in each episode, which makes the agent have little information for updating policy. Although the intrinsic reward (i.e., \textit{next-state bonus}) changes the sparse reward to dense, a limitation still exists (Section \ref{sec:next-state-introduce}).

\begin{figure}
    \includegraphics[width=0.5\textwidth]{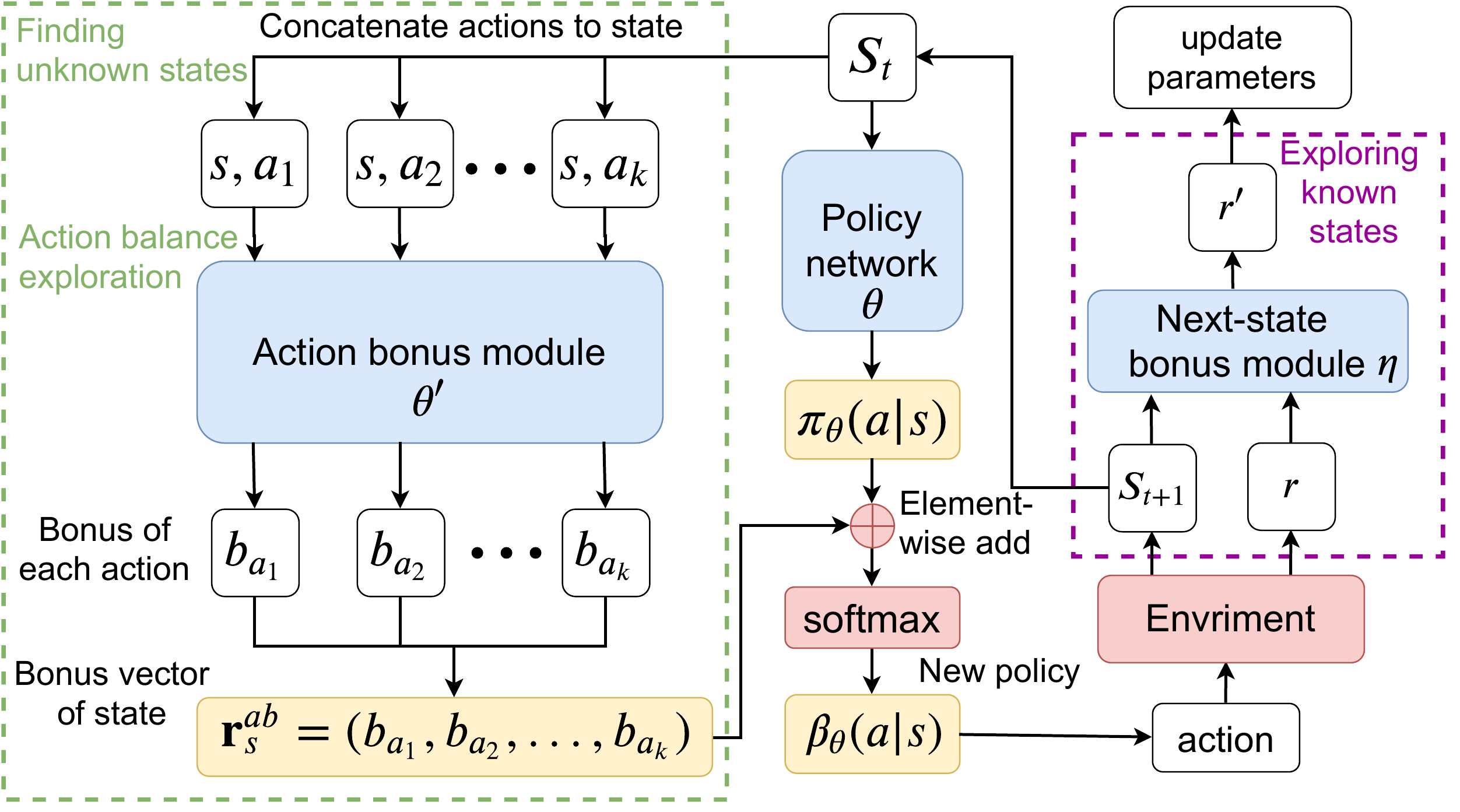}
    \caption{The overall approach of action balance RND. First, an action bonus vector is calculated with the input $s_t$. Then, the bonus vector is added to $\pi_\theta(a|s)$ to generate a new policy $\beta_\theta(a|s)$. At last, use the action sampling from $\beta_\theta(a|s)$ to interact with the environment and generate the next-state bonus. Details are described in Section \ref{sec:method}.}
    \label{fig:action_balance_model}
\end{figure}

In this work, we primarily focus on improving the performance of the \textit{next-state bonus} methods, which may limit the exploration ability in hard exploration environments. To accomplish this goal, we propose \textit{action balance exploration}, which committed to improving the effectiveness of finding unknown states. Moreover, it is convenient to combine the \textit{action balance exploration} with the \textit{next-state bonus} methods to preserve the advantages of both. An illustration of the overall combination is shown in Figure \ref{fig:action_balance_model}. Since we use RND exploration \cite{burda_exploration_2019} as the \textit{next-state bonus} method, we call the combined method as action balance RND. Specifically, we first use the action bonus module to generate the bonus vector $\mathbf{r}^{ab}_s$ of the current state. Then, we combine the bonus vector and the old policy $\pi_\theta(a|s)$, which is generated by the policy net, by an element-wise add. This will give us a new policy $\beta_\theta(a|s)$ that takes the frequency of selecting each action into consideration. After generating action by $a\sim\beta_\theta(a|s)$, we use it to interact with the environment and turn into the \textit{next-state bonus} module, which is an ordinary RND exploration and will generate a new bonus to modify the environment reward. Finally, all parameters are updated with the modified samples.

In the following sections, we first analyze the drawback of the \textit{next-state bonus} methods. Then, we introduce the details of \textit{action balance exploration} step by step.

\subsection{Drawback of the \textit{next-state bonus}}
\label{sec:next-state-introduce}
The main idea of the \textit{next-state bonus} method is to quantify the novelty of experienced states and encourage the agent to revisit novel states more often, which have already met before. To briefly summarize this exploration process, the agent first experiences some states by performing actions, which generates transition tuple $(s_t, a_t, r_t, s_{t+1})$, where $s_t$ is the state of time-step $t$, $a_t$ is the action performed at $t$, and $r_t$ is the reward given by the environment after preforming action $a_t$. Then, to encourage taking action $a_t$ when facing $s_t$, an intrinsic reward (bonus) $r_t^i$ that quantifies the novelty of $s_{t+1}$ is being calculated by taking $s_{t+1}$ as input, which gives a new transition $(s_t, a_t, r_t, s_{t+1}, r_t^i)$. At last, the exploration bonus $r_t^i$ is combined with the environment reward $r_t$ (e.g., linear combination) and affects the action selection in future steps by gradient-based methods \cite{pathak_curiosity-driven_2017, burda_exploration_2019, savinov_episodic_2019, taiga_benchmarking_2019}. Specifically, a novel next state $s_{t+1}$ will give a high exploration bonus and raise the probability of choosing action $a_t$, which will result in more visits to $s_{t+1}$.

As described above, the exploration process in \textit{next-state bonus} methods is driven by the next state $s_{t+1}$ and take effect by generating additional reward (bonus), which means the agent must obtain the next state first in order to generate exploration bonus. Besides, the bonus, which will be combined with the environment reward, only encourage to perform the specific action to reach $s_{t+1}$. In the case that an agent is fully dominated by the \textit{next-state bonus} during exploration and does not use any other exploration strategies, including $\epsilon-greedy$ and sampling action from policy distribution, the agent will always follow the most novel states, which possess the highest bonus, that have already visited before. In other words, the \textit{next-state bonus} methods force the agent to follow existing experiences for deep exploration but have little effect on finding states that never experienced. This exploration phenomenon may lead the agent to pay overmuch attention in exploring known states, but ignore finding unknown states, which is also important for exploration.

\subsection{Random network distillation module}
\label{sec:rndm}

In this section, we introduce a module used in our \textit{action balance exploration} method, which is called \textit{random network distillation module}. This module is proposed in RND exploration \cite{burda_exploration_2019} and is used to measure the occurrence frequency of input which is continuous or not enumerable.

\textit{Random network distillation module} transforms the counting process into a supervised learning task by using two neural networks: \textit{target} network and \textit{predictor} network, which use the same architecture. The target network is fixed and randomly initialized, it generates target value by mapping the input to an embedding representation $f:\mathcal{O}\to\mathbb{R}^k$. The predictor network $\hat{f}:\mathcal{O}\to\mathbb{R}^k$ tries to predict the target value and is trained to minimize the MSE: 
\begin{equation}
    r^i_t=\|\hat{f}(s;\eta)-f(s)\|^2
    \label{eq:intrinsic_reward}
\end{equation}
Where $\hat{f}$ is parameterized by $\eta$. 

In general, based on the fact that the loss of specific input will decrease as training times increase \cite{burda_exploration_2019}, the prediction errors of novel inputs are expected to be higher. This makes the intrinsic reward $r^i_t$ establish a relationship with the occurrence frequency of input and has the ability to quantify the novelty of it.

\begin{algorithm}[t]
\SetAlgoLined
\KwIn{Initial state $s$, policy network $\pi_\theta(a|s)$, action bonus module $\hat{g}(s, a^E;\theta')$, \textit{next-state bonus} module $\hat{f}(s;\eta)$, dimension of action space $k$.}
\Repeat{Max iteration or time reached}{
    \For{i=1,...,k}{ 
        Obtain embedded representation $a^E_i$ of action $a_i$ by Eq. (\ref{eq:channel_map}) (opt)\\
        Calculate action bonus $b_{a_i}$by :\\
            {\begin{center}
            $b_{a_i} = r^{ab}(s, a_i)  = \|\hat{g}(s, a^E_i;\theta')-g(s, a^E_i)\|^2$
            \end{center}}
    }
    Obtain action bonus vector $\mathbf{r}^{ab}_s$ of state $s$ by concatenating the action bonuses:\\
      {\begin{center}
        $\mathbf{r}^{ab}_s =(b_{a_1}, b_{a_2}, ..., b_{a_k})$
      \end{center}}
    Modify the original policy by:\\
        {\begin{center}
        $\beta_\theta(a|s) = softmax(\pi_\theta(a|s) + Normalize(\mathbf{r}^{ab}_s))$
        \end{center}}
    Generate action by $a\sim\beta_\theta(a|s)$.\\
    Interact with the environment to get the next state $s_{t+1}$ and external reward $r$.\\
    Generate intrinsic reward $r_t^i$ with $s_{t+1}$ and new reward $r'$:\\
        {\begin{center}
            $r_t^i = \|\hat{f}(s_{t+1};\eta)-f(s_{t+1})\|^2$\\
            $r' = r + r_t^i$
        \end{center}}
    Collect samples $(s, a, r', s_{t+1})$.\\
    Generate policy loss $l_p$ (PPO is used here) with $(s, a, r', s_{t+1})$.\\
    Update parameter $\theta, \theta', \eta$ by minimizing the overall loss:
    {\begin{center}
    	$l_t(\theta, \theta', \eta) = l_p + \|\hat{g}(s, a^E;\theta')-g(s, a^E)\|^2 + \|\hat{f}(s_{t+1};\eta)-f(s_{t+1})\|^2$
    \end{center}}
    $s\leftarrow{s_{t+1}}$\\
}

 \caption{Action balance RND}
 \label{alg:ab_exploration}
\end{algorithm}

\subsection{Action bonus module}

\label{sec:record_action_selected_fre}
The goal of the \textit{action balance exploration} is to balance the frequency of selecting each action in each state. Thus, we need to record the occurrence frequency of the state-action pair $(s, a)$, instead of only the state $s$. To accomplish this goal in a more general way, we use the \textit{random network distillation module} to count the frequency of selecting each action. The bonus of an action $a$ at state $s$ is given by:
\begin{equation}
\begin{split}
    r^{ab}(s, a) &= \|\hat{g}(s, a^E;\theta')-g(s, a^E)\|^2 \\
  	a^E &= Embedding(a)
    \end{split}
    \label{eq:action_bonus}
\end{equation}
Where $g$ and $\hat{g}$ map the input to an embedding vector of $\mathbb{R}^k$ and have the same role as $f$ and $\hat{f}$ do respectively, $a^E$ is the fixed embedded representation of action $a$ (e.g., one-hot embedding). This bonus can be used to guide the exploration in future learning.

Another thing needs to be declared is how we process the input of the state-action pair. Since $g$ is a neural network, the most common way is using the combination of $s$ and $a$ as one input (i.e., using $a$ as additional features for $s$) and obtaining the output by one computation. However, the proportion of action features takes in this combination will directly influence the output. For example, when $a$ takes a very low proportion in the combination of $(s, a)$, the action bonus $r^{ab}(s, a)$ will be dominated by $s$ and become irrelevant to $a$, vice versa. In an ideal condition, we will obtain the perfect output as we expect when $s$ and $a$ have an equal proportion in the input combination, which makes the output is decided by $s$ and $a$ equally.

\begin{figure}
    \centering
    \includegraphics[width=0.35\textwidth]{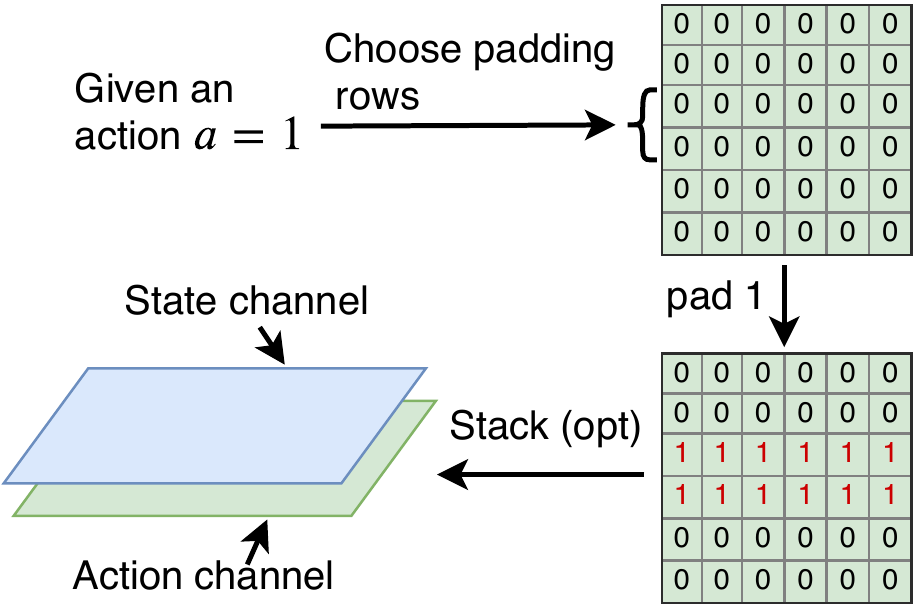}
    \caption{Illustration of mapping the 1-d action index to a 2-d array. Given an action with index 1, first, choose the padding rows. Then pad the selected rows with a specified value (e.g., 1) to get the 2-d representation of input action. This array can be regarded as another input channel for convolution architectures.}
    \label{fig:action_channel}
\end{figure}

Although the commonly used one-hot encoding is more recognizable than just the index of action, the one-hot vector may not suitable when the dimension of the state is much higher than the dimension of the encoded action. Based on this situation, we propose an encoding method that maps 1-d action to a 2-d array, which is suitable for 2-d states. Specifically, given a default $m\times{n}$ zero matrix $\mathbf{M}\in\mathbb{R}^{m,n}$, the action is represented by:
\begin{equation}
	\begin{split}
    &\mathbf{M}_{i,\ast}=c,\\
    &\forall i \in \{a\times{\lfloor\frac{m}{k}}{\rfloor} , a\times{\lfloor\frac{m}{k}}{\rfloor}+1, ..., (a+1)\times{\lfloor\frac{m}{k}}{\rfloor} -1 \}
    \end{split}
    \label{eq:channel_map}
\end{equation}
Where $a$ is the index of action, $k$ is the dimension of action space and $c$ is the padding value. The rows of $\mathbf{M}$ are divided into ${\lfloor\frac{m}{k}}{\rfloor}$ parts and a specific part is padded with $c$ according to action index $a$ (Figure \ref{fig:action_channel}). Since this 2-d array can be regarded as another input channel for convolution neural networks, we call it \textit{action channel}.

\subsection{Applying action bonus on exploration}
\label{sec:applying_action_frequency_exploration}
Given a state \textit{s}, we can obtain the frequency of selecting each action by using $r^{ab}(s, a)$. Then, we concatenate the bonus of each action to obtain the bonus vector of state $s$:
\begin{equation}
	\begin{split}
  		\mathbf{r}^{ab}_s &=(b_{a_1}, b_{a_2}, ..., b_{a_{k}}) \\
        &= (r^{ab}(s, {a_1}), r^{ab}(s, {a_2}), ..., r^{ab}(s, {a_k}))
  	\end{split}
\end{equation}

Before using the bonus vector $\mathbf{r}^{ab}_s$ to influence the exploration, normalization is performed. On the one hand, normalization raises the difference between each element in $\mathbf{r}^{ab}_s$, whose values are small, by scaling them according to a baseline, which means more straightforward to encourage or restrain on each action. On the other hand, this modification is harmless since it does not either change the relative relation of elements in $\mathbf{r}^{ab}_s$ nor disturb the outputs of other inputs. Then, this modified bonus vector will directly add to the original policy $\pi_\theta(a|s)$:
\begin{equation}
  \beta_\theta(a|s) = softmax(\pi_\theta(a|s) + Normalize(\mathbf{r}^{ab}_s))
\end{equation}
At this point, we obtain a new policy $\beta_\theta(a|s)$ which considers the frequency of selecting each action and the behavior action will be sampled from policy $a\sim\beta_\theta(a|s)$. 

Note that, since the bonus vector $\mathbf{r}^{ab}_s$ is calculated before actually take one action, we can use all actions as input to calculate the bonus vector $\mathbf{r}^{ab}_s$. Besides, the behavior policy $\beta_\theta(a|s)$ is slightly different from the target policy $\pi_\theta(a|s)$. Theoretically, it makes the action balance RND become an off-policy method and needs correction. We find the method works well without any correction in our experiments.

\subsection{Next-state bonus and parameters update}
\label{sec:parameter_update}

After obtaining the action to be performed from $\beta_\theta(a|s)$ in Section \ref{sec:applying_action_frequency_exploration}, the following process is just the same as an ordinary RND exploration \cite{burda_exploration_2019}. We first interact with the environment by action $a$ to get the next state $s_{t+1}$. Then, the intrinsic reward $r_t^i$ of $s_{t+1}$ is calculated by Eq. (\ref{eq:intrinsic_reward}) and a modified reward $r'$ is calculated by linear combination: $r' = r + r_t^i$, which gives a new transition $(s, a, r', s_{t+1})$. Finally, the overall loss is calculated by:
\begin{equation}
    \begin{split}
      l_t(\theta, \theta', \eta) = l_p &+ \|\hat{g}(s, a^E;\theta')-g(s, a^E)\|^2 \\
      &+ \|\hat{f}(s_{t+1};\eta)-f(s_{t+1})\|^2
      \label{eq:overall_loss}
    \end{split}
\end{equation}
Where the first item is the policy loss of specifying algorithm (e.g., PPO), the second and third are the prediction errors of the action bonus module and the \textit{next-state bonus} module respectively. The pseudocode of the action balance RND is presented in Algorithm \ref{alg:ab_exploration}. 

\section{EXPERIMENTS}
The primary purpose of our experiments is to demonstrate that the \textit{action balance exploration} has a better performance in finding unknown states and can speed up the exploration process of the \textit{next-state bonus} methods in some environments. Thus, we compare three exploration methods: random, RND exploration (one implementation of the \textit{next-state bonus} method), and our action balance RND (the combination of our \textit{action balance exploration} and the \textit{next-state bonus} RND). We first compare the ability to find unknown states in a grid world that is the complete absence of rewards. Second, we compare the exploratory behaviors of different methods including the \textit{action balance exploration}, which removes the \textit{next-state bonus} (RND) from the action balance RND and the agent is completely guided by the action bonuses. Third, to demonstrate how much the differences in finding unknown states will influence actual tasks, we construct another environment of reaching goals in grid word. Specifically, to make the experiment results more accurate, we use a fixed-function to calculate the intrinsic reward of each grid in RND exploration since the state is enumerable, instead of a neural network.

Finally, we test our method on six hard exploration video games of Atari. In this experiment, an ordinary RND that using a neural network to generate intrinsic reward is applied. The code is published here\footnote{https://github.com/NeteaseFuxiRL/action-balance-exploration}.


\subsection{Comparison of finding unknown states}
\label{sec:find_new_state}

\begin{figure}
\centering
    \begin{subfigure}{0.236\textwidth}
        \centering
        \includegraphics[ width=\textwidth]{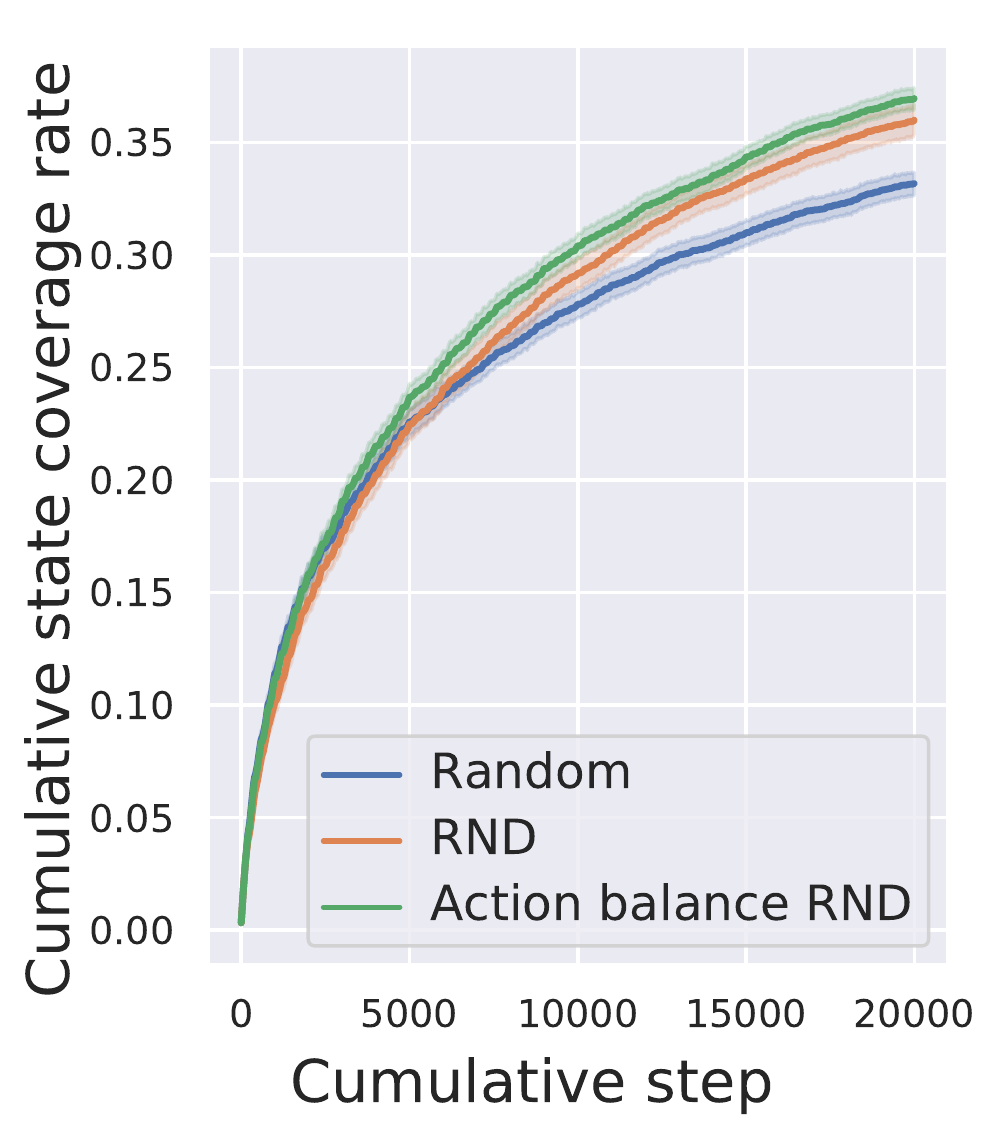}
        \caption{State coverage rate}
        \label{fig:state_visited_rate}
    \end{subfigure}
    \hfill
    \begin{subfigure}{0.236\textwidth}
		\centering
        \includegraphics[ width=\textwidth]{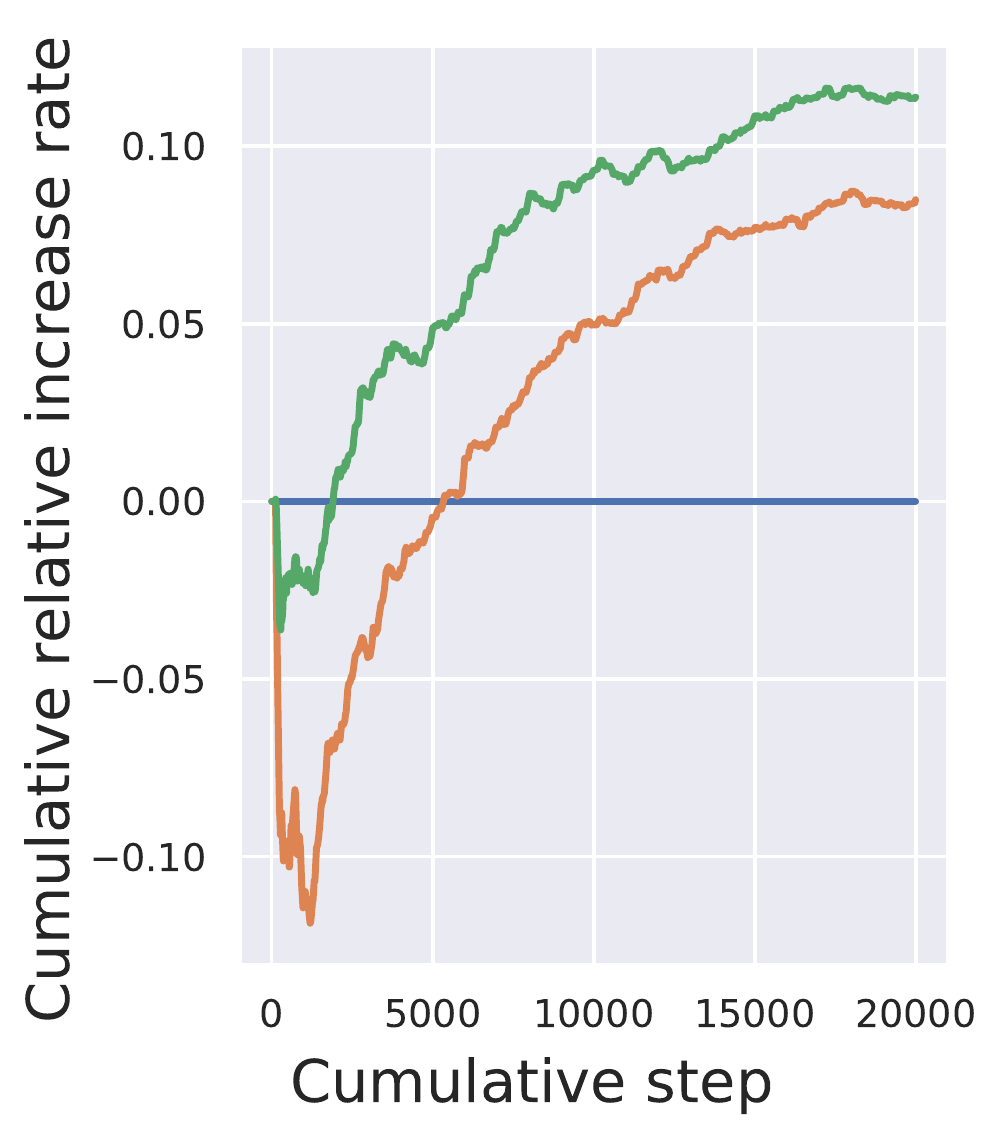}
        \caption{Relative increase rate}
        \label{fig:state_visited_rate_percentage_increase}
     \end{subfigure}
  \caption{Comparison of finding unknown states in a no-ends grid world with multiple episodes. The axes show the cumulative values.}
  \label{fig:no-ends_gridworld}
\end{figure}

\begin{figure*}
    \centering
    \begin{subfigure}{0.49\textwidth}
        \centering
        \includegraphics[ width=\textwidth]{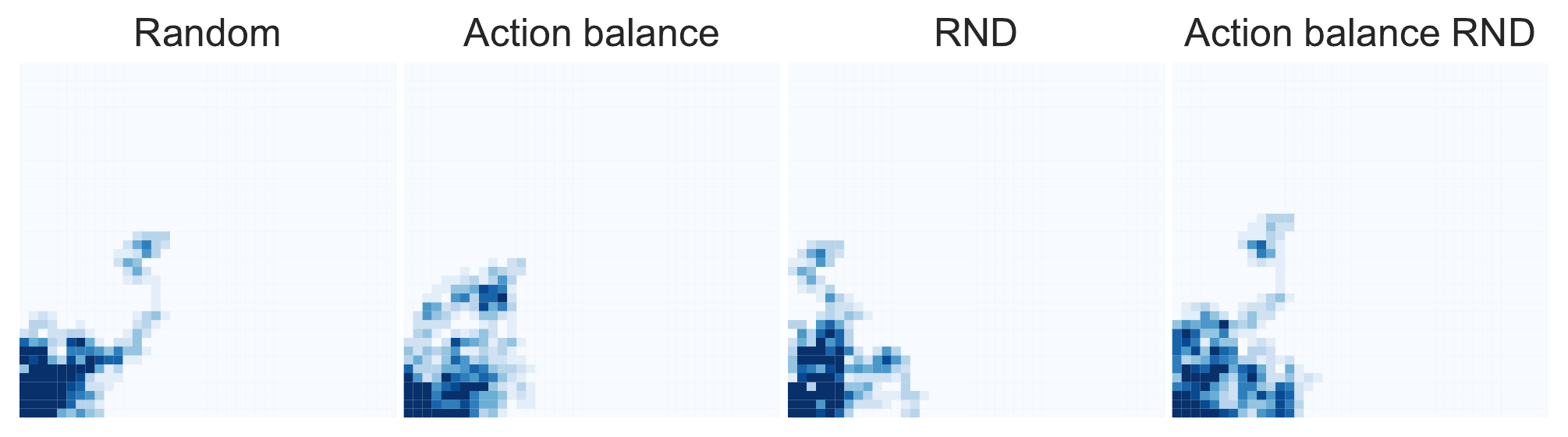}
        \caption{$step=1000$}
        \label{fig:nes100}
    \end{subfigure}
    \hfill
    \begin{subfigure}{0.49\textwidth}
        \centering
        \includegraphics[ width=\textwidth]{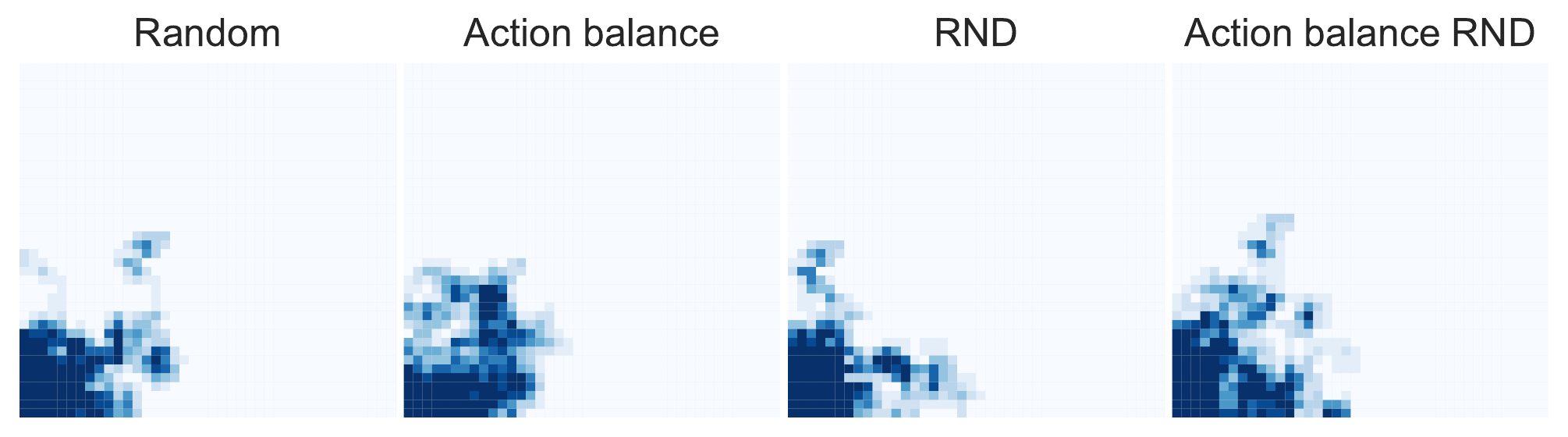}
        \caption{$step=2000$}
        \label{fig:nes200}
    \end{subfigure}
    
    \begin{subfigure}{0.49\textwidth}
        \centering
        \includegraphics[ width=\textwidth]{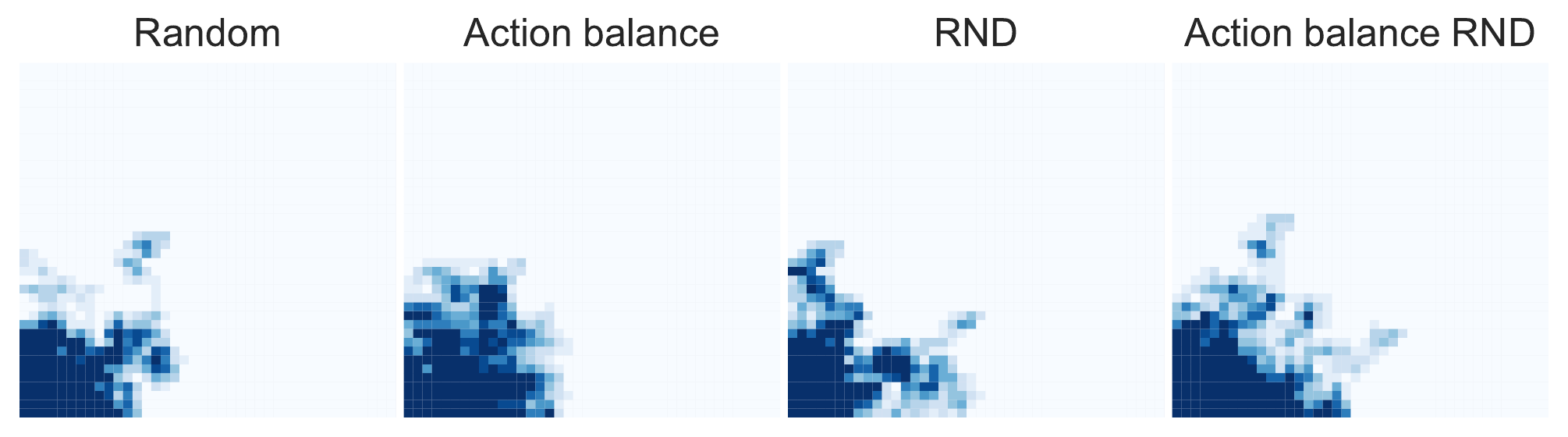}
        \caption{$step=3000$}
        \label{fig:nes300}
    \end{subfigure}
    \hfill
    \begin{subfigure}{0.49\textwidth}
        \centering
        \includegraphics[ width=\textwidth]{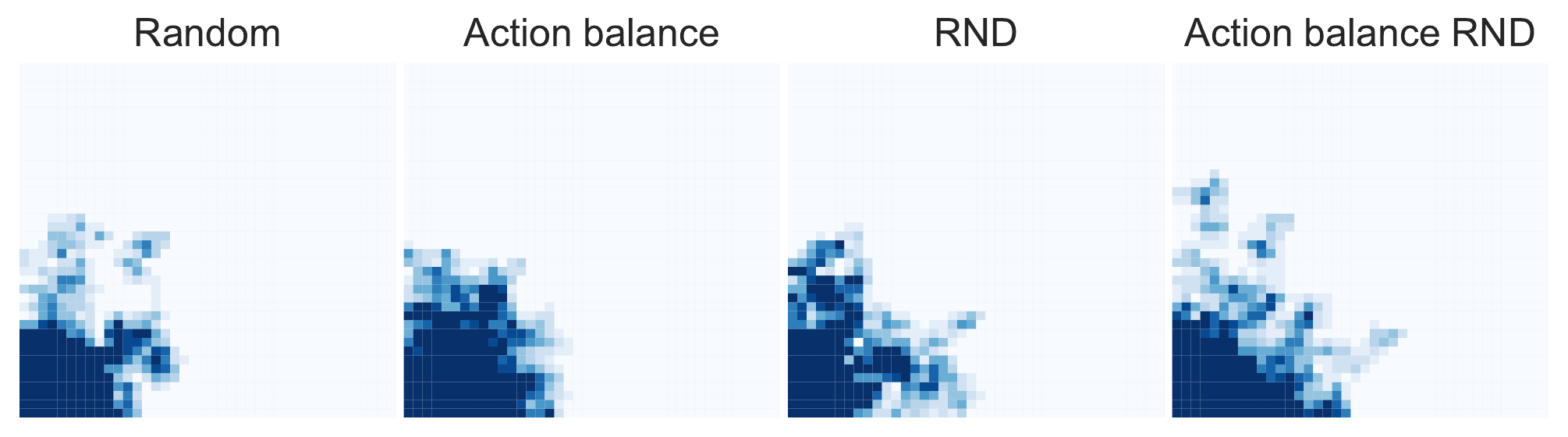}
        \caption{$step=4000$}
        \label{fig:nes400}
    \end{subfigure}

    \caption{Heat map of covered state at different steps. Darker means more visit times. From left to right at each step is random, action balance, RND, and action balance RND.}
    \label{fig:heatmap_visited_rate}
\end{figure*}

\begin{figure}
    \centering
    \begin{minipage}[t]{0.42\textwidth}
        \centering
        \includegraphics[width=\textwidth, height=0.35\textwidth]{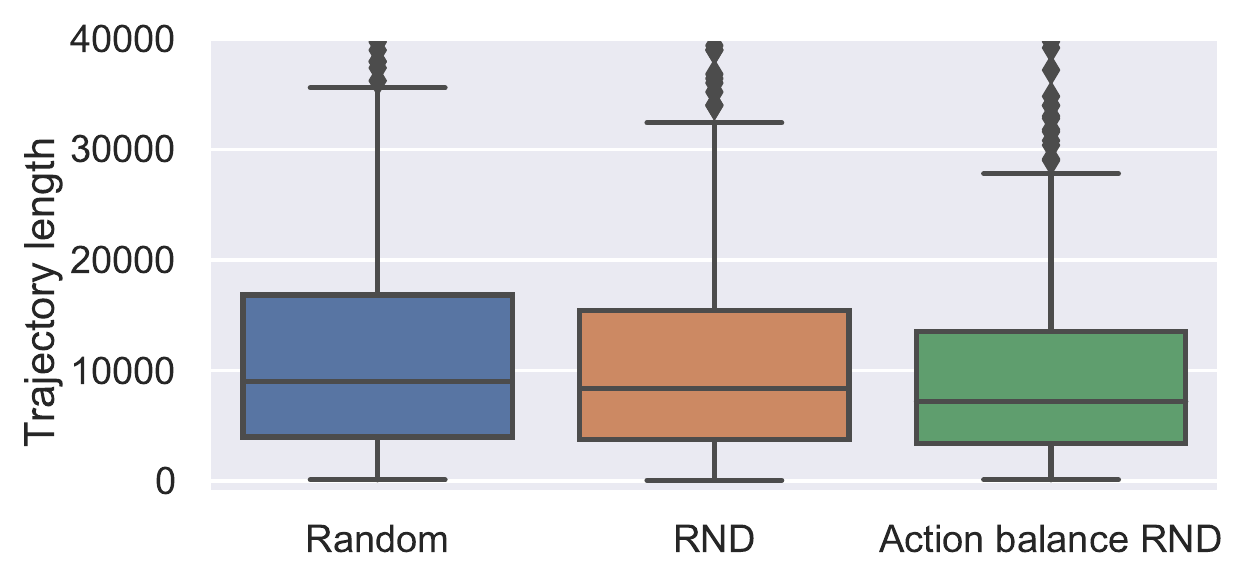}
        \caption{Trajectory length when first reaching the endpoint. Five random selected endpoints are used and each runs 100 times.}
        \label{fig:episode_length}
    \end{minipage}
    \hfill
    \begin{minipage}[t]{0.42\textwidth}
        \centering
        \includegraphics[width=\textwidth, height=0.38\textwidth]{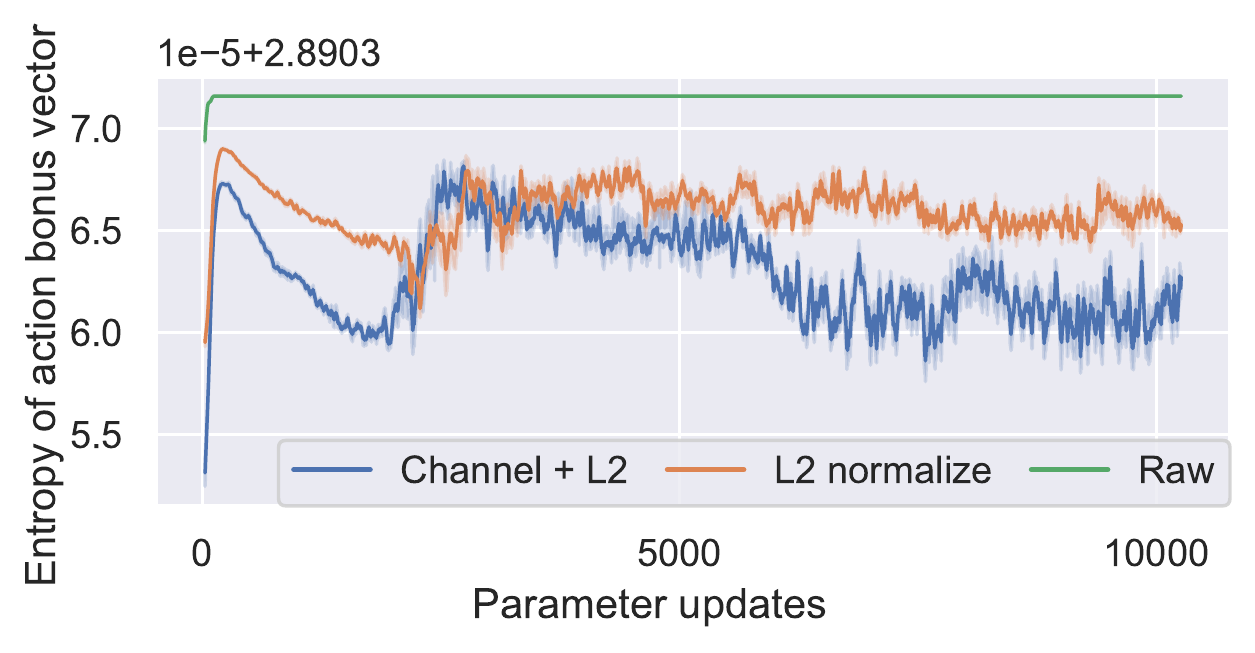}
        \caption{Entropy of action bonus vector as a function of parameter updates. Lower is better. \textit{Raw} using the original output as the bonus. \textit{L2 normalize} applies an additional L2 normalization on the raw output. \textit{Channel+L2 } adds another action channel to the input state.}
        \label{fig:action_entropy}
    \end{minipage}
\end{figure}

In this experiment, we use a simple grid world with four actions: up, down, left, right. Especially, no rewards exist in any of the grids and also no goals. What the agent could do is just walk around and explore the environment until reaching its max episode length. These settings make the grid world similar to a hard exploration environment and eliminate all the factors that may influence the exploration strategy. Based on these conditions, we can compare the differences in behaviors between exploration methods on it. Moreover, we use state coverage rate $R_s=N_{visited}/N_{all}$ to quantify the performance of exploration, where $N_{visited}$ is the number of unique states that have been visited and $N_{all}$ is the number of total unique states. A better strategy is expected to have a higher state coverage rate during the exploration process.

Specifically, the size of the gridworld is set to $40\times40$ and we use the coordinates $[x, y]$ of the agent as state representation. The start point is fixed at $P_{0}(0, 0)$ and the max episode length is 200. In order to trace the exploration process in the long term, we run 100 episodes for each method and record the state visited rates every 10 steps during this process. The results are shown in Figure \ref{fig:no-ends_gridworld}, which are averaged by 100 runs.

\textbf{Comparison of state coverage rate.} Figure \ref{fig:state_visited_rate} shows the state coverage rate along with the cumulative step number. The action balance RND covers about 15 more grids than RND at last. In order to show the relations more clearly, we calculate the relative increase rate to random (shown in Figure \ref{fig:state_visited_rate_percentage_increase}). Specifically, it is calculated by $(x-random)/random$, where $x$ is RND (i.e., state visited rate of RND) or action balance RND. As shown in Figure \ref{fig:state_visited_rate_percentage_increase}, the action balance RND overcomes the random baseline with about 2.5 times faster and outperforms RND all through, which is about 3 percent higher in the end. The results demonstrate that the action balance RND has a better ability in finding unknown states than RND, which owe to the usage of \textit{action balance exploration}.


\textbf{Analyses of exploratory behaviors.} Another result worth to say is random performs better than RND and action balance RND in the initial phase (Figure \ref{fig:state_visited_rate_percentage_increase}). This is due to the difficulty of reaching a grid increase when the random agent tries to get far from the start point. In the beginning, since most of the grids near the start point are not ever being visited, the state coverage rate rises rapidly even when using random selection. As time goes by, it becomes harder and harder for a random agent to find unknown states since most of the grids near the start point have already been visited and it is difficult to get farther for the random agent.

In contrast to RND, instead of finding unknown states, the agent tends to do deep exploration in known states, which leads to a lower coverage rate in the initial phase. However, RND makes it possible for the agent to directly go to states far from the start point and begin exploration there. Since it is much wider away from the start point, the agent has more opportunity to see unknown states in there, which results in a higher visited rate in the latter phase. After introducing the \textit{action balance exploration} into RND (i.e., action balance RND), the speed of finding unknown states has been accelerated a lot and the final visited rate is increased too.

Figure \ref{fig:heatmap_visited_rate} is the heat maps of the exploration process, which shows the exploration tendency. A random agent tends to hover around start point and generate disorder trajectories. The \textit{action balance exploration}, which is completely guided by the action bonuses, appears more serried around the start point since the agent tends to choose low-frequency actions. As for RND (i.e., \textit{next-state bonus}), although the agent succeeds in exploring a wider range, but due to the lack of strategy on finding unknown states, the agent unremittingly explores individual regions and earns nearly the same performance as random selection. As for action balance RND, the agent not only explores wider (via RND) but also more even in each direction (via the \textit{action balance exploration}) than the others. More results can be found in this video\footnote{https://youtu.be/A6VDYiJ8phc}.

\subsection{Influence in the task of reaching goals.}

\begin{table}
    \centering
    \caption{Trajectory lengths when first reach the endpoint.}
    \label{tab:episode_length}
    \begin{tabular}{cccc}
    \toprule
    end (x, y): & random & RND & action balance RND\\
    \midrule
    (0, 20) & 10583.86 & 10304.23 & \textbf{8118.19}\\
    (20, 0) & 11142.43 & 11062.97 & \textbf{8360.99}\\
    (10, 20) & 12772.29 & 13434.67 & \textbf{9248.96}\\
    (16, 16) & 18755.5 & 12339.32 & \textbf{11364.08}\\
    (20, 10) & 13464.31 & 11197.49 & \textbf{10384.51}\\
    average & 13343.678 & 11667.736 & \textbf{9495.346}\\
    
    \bottomrule
    \end{tabular}
\end{table}

\begin{figure*}
    \centering
    \includegraphics[width=0.9\textwidth]{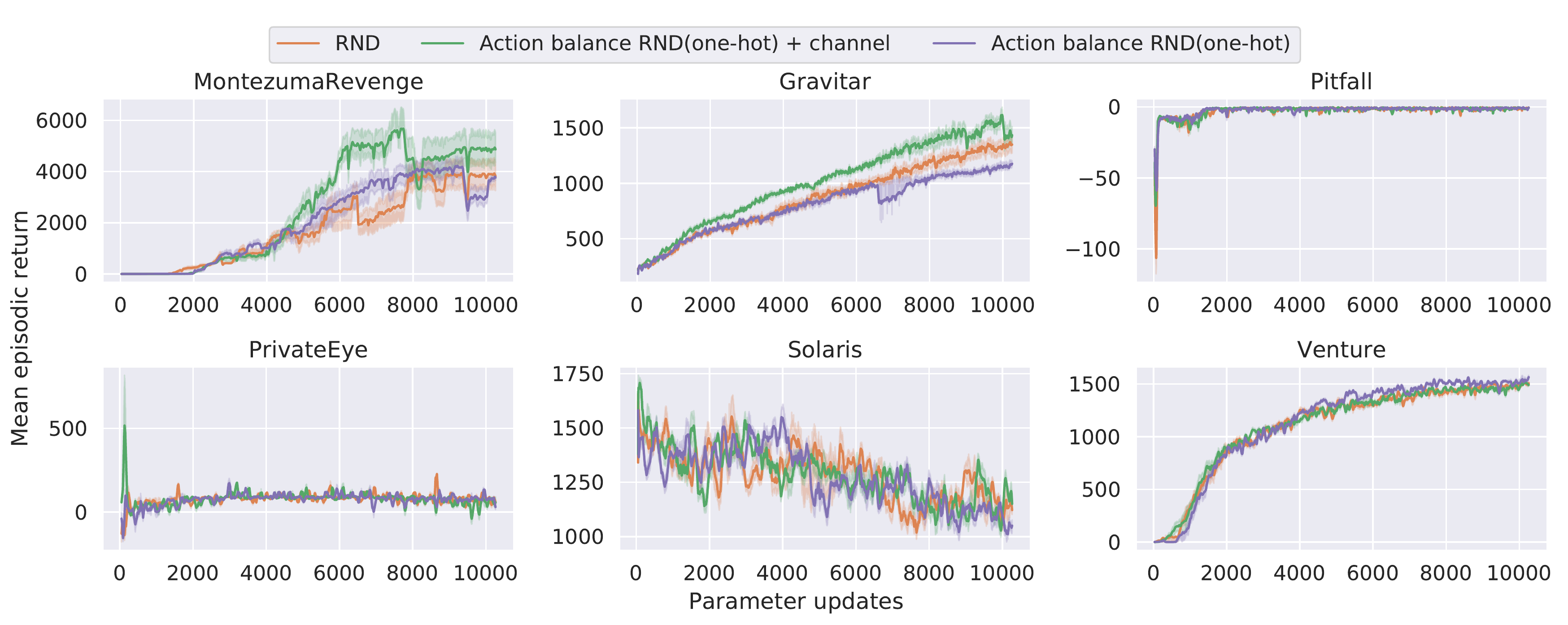}
     \caption{Mean episodic return as a function of parameter updates.}
    \label{fig:atari_reward}
\end{figure*}

This is another toy experiment that aims to show how much the differences in finding new states will influence actual tasks. We follow the fundamental settings described in section \ref{sec:find_new_state}, except there is a goal in the environment. As we only care about the exploration efficiency, we let the game finish as soon as the agent first reaches the endpoint and record the length of this trajectory for comparison. The trajectory length is expected to be smaller for a better exploration method.

\begin{table}
    \centering
    \caption{Final episodic return.}
    \label{tab:final_episodic_return}
    \begin{tabularx}{\linewidth}{>{\hsize=.5\hsize}X>{\hsize=.2\hsize}X>{\hsize=.3\hsize}X>{\hsize=.35\hsize}X}
    \toprule
    & RND & Action balance RND & Action balance RND + channel\\
    \midrule
    Montezuma's Revenge & 3812 & 3762 & \textbf{4864}\\
    Gravitar & 1349 & 1177 & \textbf{1434}\\
    Pitfall! & 0 & 0 & -1\\
    Private Eye & \textbf{59} & 30 & 57\\
    Solaris & 1122 & 1051 & \textbf{1150}\\
    Venture & 1506 & \textbf{1564} & 1491\\
    \bottomrule
    \end{tabularx}
\end{table}

In this experiment, we select 5 endpoints around the start position symmetrically to test on and run 100 times for each endpoint. As shown in Figure \ref{fig:episode_length}, random gains the highest median and third quartile number of trajectory length, RND is slightly lower than random and action balance RND obtains the lowest number. Detail results are shown in Table \ref{tab:episode_length}. The result shows that the action balance RND is about 1.23 times smaller on average than RND and beats the other two at each endpoint, which means the action balance RND can find useful information faster than RND.

\subsection{Atari}

In this section, we test our method on six hard exploration video games of Atari: Gravitar, Montezuma's Revenge, Pitfall!, Private Eye, Solaris, and Venture. First, we compare different processing methods on the action bonus vector, which tries to shape the elements in a bonus vector more distinguishable. After that is a comparison of different action embedding methods, which aims to find an appropriate representation of action. These two experiments are trying to increase the performance of \textit{action balance exploration} in different ways. Finally, we contrast the reward curves of the learning process. The mainly hyper-parameters are the same as RND and all the results are averaged by 5 runs.


\textbf{Adding action channel.} This experiment compares different action embedding methods and tries to find an appropriate representation for Atari games. Specifically, unless particularly stated, we use one-hot embedding as the default method to embed actions. Since the input state of Atari games is a 2-dimensional array, one-hot action can not directly combine to the input when calculating the action bonus. One solution is using convolution networks to process the input and generate a 1-dimensional feature, which will be concatenated to the one-hot action. We adopt this method in relevant experiments. However, the feature generated by convolution networks usually has an overlarge dimension than the one-hot action in Atari games (thousands to tens in our case), which results in too large proportion in the combination of state-action and may cover up the effect of action when calculating the action bonus of given state.

In order to increase the proportion of action when calculating the action bonus, we add another action channel to the input state (described in Section \ref{sec:record_action_selected_fre}). The dimension of the action channel is the same as the input state ($84\times84$). We use 0.01 as the padding value of the channel ($c=0.01, m=84$ in eq. (\ref{eq:channel_map})). As the result of Montezuma's Revenge shown in Figure \ref{fig:action_entropy}, using an action channel makes the entropy lower than others, which means the effect of action in calculating the action bonus is enhanced.



\textbf{Reward curves during training.} Figure \ref{fig:atari_reward} shows the mean episodic return of different games during training. Table \ref{tab:final_episodic_return} is the final returns. As we can see, the action balance RND not only earns the highest final return but also escalates much faster than RND in Montezuma's Revenge and Gravitar, and obtain almost the same results as RND in other games, which means the \textit{action balance exploration} does not bring negative effects to the exploration process at least and may be beneficial to some environments. Besides, using an additional action channel makes the results even better than only using one-hot action in most of the games.


As described in Section \ref{sec:find_new_state}, the \textit{action balance exploration} makes an agent explore the environment in a more uniform way. The rewards in Montezuma's Revenge and Gravitar are relatively uniform and dense in state space than the others, which is beneficial for the \textit{action balance exploration}. Specifically, the agent can obtain rewards by touch something in most rooms of Montezuma's Revenge. Although one needs to fire and destroy enemies in Gravitar to obtain rewards, the enemy is quite a bit in each room and it's meaningful to travel around. In contrast, positive rewards appear only in very low-frequency situations in Private Eye. Pitfall! not only suffers the lack of positive rewards but also exists negative rewards all around. Solaris is a little complex to obtain rewards than others. Venture has similar settings as Gravitar, but it is more dangerous and cannot get rewards before finding the treasure in rooms. These game settings let only a few actions (or strategies) related to the reward, which is not conducive for the \textit{action balance exploration} to take an effect.



\section{CONCLUSION}

In this work we propose an exploration method, we call \textit{action balance exploration}, which focuses on finding unknown states, contrasts to the \textit{next-state bonus} methods which aim to explore known states. We also propose a novel exploration method, action balance RND, which combines our \textit{action balance exploration} and RND exploration. The experiments on grid word and Atari games demonstrate the \textit{action balance exploration} has a better capability in finding unknown states and can improve the real performance of RND in some hard exploration environments. In the future, we want to try more complicated methods to combine the \textit{action balance exploration} and RND exploration, like adaptation coefficient, hierarchical exploration.

\bibliographystyle{IEEEtran}  
\bibliography{IEEEabrv,ab}  

\end{document}